\documentclass[final]{elsarticle}

\usepackage{hyperref}
\usepackage{csquotes}
\usepackage{amsfonts}
\usepackage{amsmath}
\usepackage{danudefs}
\usepackage{ctable}
\usepackage{tabularx}
\usepackage{flexisym}
\usepackage{graphicx}
\usepackage{caption}
\usepackage{subcaption}

\graphicspath{{./figures/}}

\usepackage{color}

\begin{document}

\begin{frontmatter}

\title{Compositional Approaches for Representing Relations Between Words: A Comparative Study}


\author{Huda Hakami\corref{mycorrespondingauthor}}

\cortext[mycorrespondingauthor]{Corresponding author}
\ead{h.a.hakami@liv.ac.uk}
\author{Danushka Bollegala}
\ead{danushka.bollegala@liv.ac.uk}
\address{Department of Computer Science, The University of Liverpool, L69 3BX, UK} 

\begin{abstract}
Identifying the relations that exist between words (or entities) is important for various natural language processing tasks such as, relational search, noun-modifier classification and analogy detection. A popular approach to represent the relations between a pair of words is to extract the patterns in which the words co-occur with from a corpus, and assign each word-pair a vector of pattern frequencies. 
Despite the simplicity of this approach, it suffers from data sparseness, information scalability and linguistic creativity as the model is unable to handle previously unseen word pairs in a corpus. In contrast, a compositional approach for representing relations between words overcomes these issues by using the attributes of each individual word to indirectly compose a representation for the common relations that hold between the two words. This study aims to compare different operations for creating relation representations from word-level representations. We investigate the performance of the compositional methods by measuring the relational similarities using several benchmark datasets for word analogy. Moreover, we evaluate the different relation representations in a knowledge base completion task.
\end{abstract}


\end{frontmatter}

\section{Introduction}
Different kinds of semantic relations exist between words such as synonymy, antonymy, meronymy, hypernymy, etc. 
Identifying the semantic relations between words (or entities) is important for various Natural Language Processing (NLP) tasks such as  knowledge base completion \citep{socher2013reasoning}, relational information retrieval \citep{Duc:WI:2010} and analogical reasoning \citep{turney2005corpus}.
To answer analogical questions of the form \enquote {\emph{a} is to \emph{b} as \emph{c} is to \emph{d} }, the relationship between the two words in each pair $(a,b)$ and $(c,d)$ must be identified and compared. For example, $(lion, cat)$ is relationally analogous to $(ostrich, bird)$ because a \emph{lion} is a large \emph{cat} as an \emph{ostrich} is a large \emph{bird}. In relational information retrieval, given the query \emph{a is to b as c is to?}
we would like to retrieve entities that have a semantic relationship with $c$ similar to that between $a$ and $b$.
For example, given the relational search query \emph{Bill Gates} is to \emph{Microsoft} as \emph{Steve Jobs} is to?,
a relational search engine~\cite{Duc:AAAI:2011} is expected to return the result \emph{Apple Inc}. 

A popular approach for representing the relations that exist between pairs of words is to extract the lexical patterns in which the pairs of words co-occur in some context \citep{turney2005corpus,turney2006similarity,bollegala2008sits}. In a text corpus, relationships between words are categorised by the patterns in which they co-occur, for instance \enquote {\emph{a} is a \emph{b}} or \enquote {\emph{b} such as \emph{a} } patterns indicate that \emph{a} is a hyponym of \emph{b}.
Following the Vector Space Model (VSM) \citep{Turney:JAIR:2010}, each pair of words is represented using a vector of pattern frequencies where the elements correspond to the number of times the two words in a given pair co-occur with a particular pattern. This representation allows us to measure the relational similarity between two given pairs of words by the cosine of the angle between the corresponding pattern-frequency vectors. We call this approach the \emph{holistic} approach, because the pairs of words are treated as a whole rather than individually. This method achieved human-level performance for measuring relational similarity on Scholastic Aptitude Test multiple-choice word analogy questions.
The average score reported for the US college applicants is 57.0\%, whereas Latent Relational Analysis (LRA), a state-of-the-art algorithm 
for measuring relational similarity using the holistic approach, obtained a score of 56.1\% \citep{turney2006similarity,turney2005measuring}.   

Despite the \emph{holistic} method achieving human-level performance, especially for relational similarity prediction tasks, a major drawback 
of the holistic approach is the data sparseness. Most of the elements in pair-pattern vector space have zero occurrences, because most related words co-occur only with a small fraction of the extracted patterns. Moreover, not every related word pair co-occur even in a large corpus. Therefore the relations that exist between words that co-occur rarely cannot be adequately represented. Another limitation of this approach is its scalability, as we must consider co-occurrences between patterns and all pairs of words. The number of all pair-wise combinations between words grows quadratically with the number of words in the vocabulary. Therefore, it is computationally costly, especially if the vocabulary size is very large ($>10^{6}$) and new words are continuously 
proposed because for each new word, we must pair it with existing words in the vocabulary. Furthermore, a continuously increasing set of patterns is required in order to cover the relations that exist between the two words in each of those word-pairs.

To overcome the above mentioned issues in the holistic approach, an alternative method that does not rely on pair-pattern co-occurrences is required. 
Such alternative methods must be able to represent the semantic relations that exist between all possible pairings of words,
requiring only semantic representations for the constituent words.
In this paper, we call such approaches for representing the relationship between words as \emph{compositional} approaches,
 because the way in which the relation representation is composed using the semantic representations of the constituent words. 
Different approaches have been proposed in the NLP community for representing the meaning of individual words based on 
the \emph{distributional hypothesis}~\citep{Harris:1985}, which states that the meaning of a word can be predicted by the words that co-occur with it in different contexts.
Counting-based approaches~\citep{baroni-dinu-kruszewski:2014:P14-1} represent the meaning of a word by a potentially high-dimensional sparse vector, where each dimension corresponds to a particular
word that co-occurs with the word under consideration in some context. The values of the dimensions are computed using some word association measure
such as the pointwise mutual information or log-likelihood ratio~\citep{turney2010frequency}.

Prediction-based approaches have also been used for representing the meanings of words using vectors~\citep{Milkov:2013,Pennington:EMNLP:2014}. Instead of counting the co-occurrences of a target word in its context, Neural Network Language Model (NNLM) \citep{bengio2003neural} uses distributional information in a corpus to maximise the probability of predicting the target word from the surrounding context. This procedure embeds the words into a low-dimensional latent dense vector space model. \citet{mikolov2013linguistic} show that the learnt word embeddings using recurrent neural network language model \citep{mikolov2010recurrent} captures linguistic regularities by simply applying vector offset and addition operators. They evaluate the accuracy of
the  learnt word representation by applying them to solve word analogy questions of the form \enquote {\emph{a} is to \emph{b} as \emph{c} is to \emph{d} }, where \emph{d} is unknown and it is typically selected from a subset of words from the vocabulary such that  $\vec{v}_b-\vec{v}_a+\vec{v}_c$ $\approx$ $\vec{v}_d$ (we denote the vector representing the word \emph{a} as $\vec{v}_a$). 
Arguably, one of the most popular examples is the following:  $\vec{v}_{king}-\vec{v}_{man}+\vec{v}_{woman}$ $\approx$ $\vec{v}_{queen}$, which describes a gender relationship. 

In compositional approaches, the meaning of longer lexical units such as phrases or sentences are composed by applying some operators on the semantic representations for individual words. The \emph{principle of compositionality} states that the meaning of the hole is a function of the meaning of the parts \cite{Frege:1892}. Over the years, researchers in compositional semantics have applied different compositional approaches to extend the meaning of individual words to larger linguistic units \cite{Mitchell:COGSCI:2010,Baroni:EMNLP:2010,Guevara:GEMS:2010}. 
However, the problem of representing the meaning of a sentence differs from our problem, representing the relation between two words, in several important ways. First, a sentence would often contain more than two words, whereas we consider word pairs that always contain exactly two words. Second, a good sentence representation must encode the meaning of the sentence in its entirety, ideally capturing the meanings of salient content words in the sentence. On the other hand, in relation representation, we are \emph{not} interested in the meanings of individual words, but the relationship between the two words in a word pair. For example, given the word pair (\emph{ostrich}, \emph{bird}), the semantics associated with \emph{ostrich} or \emph{bird} is not of interest to us. We would like to represent the relation \emph{is-a-large} that holds between the two words in this example. It is true that most of the compositional operators that have been proposed in prior work on sentence representations such as vector addition or element-wise multiplication could be used to create relation representations for word pairs, but there is no guarantee that the exact same operators that have found to be effective for sentence representation will be accurate for relation representation. As we see later in our experiments, vector offset, which does not scale up to sentences turns out to be a better operator for relation representation.

In this paper, we explore several compositional approaches for
creating representations for the relations between words. In brief, we need a function that takes two vector representations for each word in a given word-pair to generate a vector for the relation that exists between those words. Our contributions in this work can be summarised as follows:
\begin{itemize}
\item An empirical comparison of the unsupervised compositional operators (offset, concatenation, addition and element-wise multiplication) to represent relations between words.
\item Investigate the performance of those operators on relational similarity and a relational classification tasks using five different word-analogy benchmark datasets. 
\item Evaluate such operators on a  knowledge base completion task.
\item Understand to what extent the performance of those methods change across different word representation methods including counting-based and predicting-based approaches. 
\item Systematically examine how the performance of different compositional operators are affected by the dimensionality of the word embeddings.
\end{itemize}

Our study shows that the offset operator for relational compositionality outperforms other compositional operators on word-analogy datasets. For knowledge base completion, element-wise multiplication shows its ability to capture relations between entity embeddings for a given knowledge graph.    


\section{Related work}
\label{sec:related}

Representing the meaning of individual words has received a wide attention in NLP. 
Different representation methods have been proposed using the distributional semantics of the words in a corpus to obtain a vector space model of semantics where each word is represented in term of its surrounding lexical contexts. The distributional hypothesis is summarised by \citet{firth1957synopsis} as follows ``You shall know a word by the company it keeps'', which means that the words that appear in similar contexts share similar meanings. 
The traditional \emph{count-based} word representations count the co-occurrences of a  word with its neighbouring words in a specific window size.
In practice however this method generates high dimensional and sparse vectors \citep{baroni2010distributional,turney2010frequency}. 

Recently, instead of counting the occurrences between words and contexts, machine learning techniques have been applied in NLP to directly learn dense words vectors by \emph{predicting} the occurrence of a word in a given context.  
For example, skip-gram and continuous bag-of-words models 
learn vectors that maximise the likelihood of co-occurrence contexts in a corpus \citep{Milkov:2013}. 
The word representations learnt via prediction-based methods are often referred to as \emph{words embeddings}
because the words are represented (embedded) using vectors in some lower-dimensional space. 
In addition to the fact that the learnt semantic space represents semantically similar words close to each other, \citet{mikolov2013linguistic} report that word embeddings capture relational information between words by simple linear offset between words vectors.
 In their study, they propose an analogical reasoning task to evaluate word embeddings. 
 To answer analogical questions of the form \enquote {\emph{a} is to \emph{b} as \emph{c} is to ? }, they subtract the embedding of word \emph{b} from \emph{a} and then add the embedding of \emph{c}. Next, a word in the entire vocabulary set that is the most similar to the generated vector is selected as the answer. They refer to this method for solving analogy as \textsf{3CosAdd}. Following this work, alternative methods have been proposed and compared with \textsf{3CosAdd} for analogical reasoning \cite{levy2014linguistic,linzen2016issues,drozdword}. These prior studies focus on proposing methods for solving word analogy problems given word embeddings
 but do not consider composing representations for the relations that exist between two words in a word-pair.
 
\citet{vylomova2015take} conduct a study to evaluate how well the offset method encodes relational information between pairs of words. 
They test the generalisation of the offset method across different types of relations by evaluating the relational vectors generated by the offset method
in an unsupervised (clustering) task and a supervised (classification) task. 
They conclude that information about syntactic and semantic relations are implicitly embedded in the offset vectors, especially under supervised learning. 
However, they find that the offset method does not capture semantic relations to the same level of accuracy as it captures the syntactic relations.  

Many compositional operators have been proposed for the purpose of representing sentences \citep{Baroni:EMNLP:2010,Mitchell:ACL:2008}. For example, \citet{Mitchell:ACL:2008} introduce additive and multiplicative models for sentence representations, whereas \citet{Nickel:AAAI:2016}
proposed circular correlation for relational composition.
However, to the best of our knowledge, there exist no work that compares different compositional operators for the purpose of relation representation.
To this end, our study aims to systematically evaluate how well the contribution of word embeddings to represent relations between words by comparing different compositional operators under unsupervised settings.  

\section{Relation Composition}
\subsection{Compositional operators}
\label{sec:operators}

Our goal in this paper is to compare different compositional operators for the purpose of composing representations for the relation between two words, given the word embeddings for those  two words.
In this work, we assume that pre-trained word embeddings are given to us, and our task is to use those word embeddings to compose relation representations.
Specifically, given a word-pair $(a, b)$, consisting of two unigrams $a$ and $b$, represented respectively by their embeddings
$\vec{v}_{a}, \vec{v}_{b} \in \R^{n}$, we propose and evaluate different compositional operators/functions
that return a vector $\vec{v}_{r}$ given by \eqref{eq:f} that represents the relationship between $a$ and $b$.
\begin{equation}
\label{eq:f}
\vec{v}_r=f(\vec{v}_a,\vec{v}_b)
\end{equation}
In this paper, we limit our study to non-parametric functions $f$. 
Parametric functions that require labelled data for computing the optimal values of the parameters for generating relation representations are beyond the
scope of this paper.  

We use the following operators to construct a vector for a given pair of words:
\begin{description}

\item {\textsf{PairDiff}: }
Pair Difference operator has been used by \citet{mikolov2013linguistic} for detecting syntactic and semantic analogies 
using the offset method. For example, given a pair of words  $(a,b)$, they argue that $(\vec{v}_b-\vec{v}_a)$ produces a vector that captures the relation that exits
 between the two words \emph{a} and \emph{b}. 
Under the PairDiff operator, a resultant relation representation vector has the same dimensionality as the input vectors. 
The PairDiff operator is defined as follows:
\begin{equation}
\label{eq:pairdiff}
\vec{v}_r= (\vec{v}_b - \vec{v}_a)
\end{equation}
PairDiff captures the information related to a semantic relation by the direction of the resultant vector.
Similar relations have shown to produce parallel vectors in prior work on word embedding learning~\citep{Pennington:EMNLP:2014}.
Such geometric regularities are useful for NLP tasks such as solving word analogies~\citep{mikolov2013linguistic}.

\item{\textsf{Concat.}:}
 The linear concatenation of two $n$-dimensional vectors 
$\vec{v}_a = {(a_1, \ldots, a_n)}^\top$ and $\vec{v}_b = {(b_1, b_2, \ldots, b_n)}^\top$
produces a $2n$-dimensional vector $\vec{v}_r $ given by,
\[\vec{v}_{r} = (a_1, a_2, \ldots, a_n, b_1, b_2, \ldots, b_n)\T .\] 
$ \vec{v}_{r}$ can then be used as a proxy for the relationship between $a$ and $b$.
Vector concatenation retains the information that exist in both input vectors in the resulting composed vector.
In particular, vector concatenation has been found to be effective for combining multiple source embeddings to a single meta embedding~\cite{Yin:ACL:2016}.
However, one disadvantage of concatenation is that it increases the dimensionality of the relation representation compared to that in the input word embeddings.

\item{\textsf{Mult}: }
We apply element-wise multiplication between $\vec{v}_a$ and  $\vec{v}_b$ such that the \emph{i$^{th}$} dimension of $\vec{v}_r$ has the value of multiplying the \emph{i$^{th}$} dimensions of the input vectors. 
Applying  element-wise multiplication generates a vector in which the dimensions common to both words receive non-zero values. \textsf{Mult} operator is defined as follows:
\begin{equation}
\begin{split}
\vec{v}_r= (\vec{v}_a \odot \vec{v}_b)\\
\vec{v}_{r_i}=\vec{v}_{a_i} \vec{v}_{b_i}
\end{split}
\end{equation}
Element-wise multiplication has the effect of selecting dimensions that are common to the embeddings of both words for representing the relationship between those words.
Prior work on compositional semantics have shown that element-wise multiplication to be an effective method for composing representations for larger lexical units such as
phrases or sentences from elementary lexical units such as words~\citep{Mitchell:ACL:2008}.
However, element-wise multiplication has an undesirable effect when the embeddings contain negative values. For example, two negative-valued dimensions can generate a positive-valued
dimension in the relational representation. If the relations are directional (asymmetric), then such changes of sign can incorrectly indicate opposite/reversed relations between words.
For example, \citet{Baroni:EMNLP:2010} report that word embeddings created via singular value decomposition performs poorly when composing phrase representations
because of this sign-flipping issue. 
As we see later in Section~\ref{ExperRes}, \textsf{Mult} also suffers from data sparseness because if at least one of the corresponding dimensions
in two word embeddings is zero (or numerically close to zero), then the resultant dimension in the composed relational vector becomes zero.
Our experimental results suggest that more than negativity, sparseness is problematic for the \textsf{Mult} operator.
However, to the best of our knowledge, the accuracy of element-wise multiplication has not been evaluated so far in the task of relation representation.

\item{\textsf{Add}:}
 We apply element-wise addition between $\vec{v}_a$ and  $\vec{v}_b$ such that the \emph{i$^{th}$} dimension of $\vec{v}_r$ has the value of adding the \emph{i$^{th}$} dimensions of the input vectors, given as follows:
 \begin{equation}
\label{eqn:add}
\begin{split}
\vec{v}_r= (\vec{v}_a + \vec{v}_b) \\
\vec{v}_{r_i}=\vec{v}_{a_i}+\vec{v}_{b_i}
\end{split}
\end{equation}
\end{description}

Element-wise multiplication and addition have been evaluated in compositional semantics for composing phrase-level or sentence-level representations from
word-level representations~\citep{mitchell-lapata:2009:EMNLP,Mitchell:ACL:2008}. 
In the context of relations, a relationship might materialise between two entities because they share many attributes in common.
For example, two people might become friends in social media because they discover they have many common interests.
Consequently, element-wise addition and multiplication emphasise such common attributes by adding their values together when composing the corresponding relation representation. 
In this work, we hypothesise that some relations are formed between entities because they have common attributes. By pairwise addition or multiplication
of the attributes of two given words, we are emphasising these common attributes in their relational representation.

Element-wise operators between word vectors assume that the dimensions of the word representation space are linearly independent. 
Alternatively, we can consider that the dimensions are cross-correlated and use cross-dimensional operators
(i.e. operators that consider $i^{th}$ and $j^{th}$ dimensions for $i = j$ as well as $i \neq j$)
instead of element-wise operators to create relation representations.
For this purpose, given a word representation matrix $\mat{W}$ $ \in \mathbb{R}^{m \times n} $ of \emph{m} words and \emph{n} dimensions, we create a correlation matrix $\mat{C}$ $ \in \mathbb{R}^{n \times n} $ in which the $\mat{C}_{ij}$ element is the \emph{Pearson} correlation coefficient of $\mat{W}_{:,i}$ and $\mat{W}_{:,j}$, (i.e., the \emph{i$^{th}$} and the \emph{j$^{th}$} dimensions for all of the represented words). 
In our preliminary experiments, for the pre-trained word embeddings we use as inputs, 
we found that the correlation coefficients between $i$, $j (\neq i)$ dimensions are close to zero,
 which indicates that the dimensions are indeed uncorrelated. Consequently, for the prediction-based word embeddings we used
 in this comparative study, we did not obtain any significant improvement in performance by using 
 cross-dimensional operators. Therefore, in the remainder of the paper, we do not consider cross-dimensional operators.

\subsection{Input Word Embeddings}
\label{WRM}

We consider three widely used prediction-based word embedding methods namely, Continuous Bag-of Words (CBOW), Skip-gram (SG)\footnote{\url{https://code.google.com/archive/p/word2vec/}}\citep{Milkov:2013} and Global Vector Prediction (GloVe)\footnote{\url{http://nlp.stanford.edu/projects/glove/}} \citep{Pennington:EMNLP:2014}. CBOW and SG models the task of learning word embeddings as predicting words that co-occur in a local
contextual window. 
The latent dimensions in the embeddings can be seen as representing various semantic concepts that are useful for representing the meanings of words.
However, unlike in counting-based word embeddings, in prediction-based word embeddings the dimensions are not explicitly associated with a particular word or a class of words.
In brief, CBOW learn word embeddings by maximising the probability of predicting a target word from the surrounding context words, whereas SG aims to predict surrounding context words given a target word in some context. On the other hand, GloVe learning method considers global co-occurrences over the entire corpus.
Specifically, GloVe first builds a co-occurrence matrix between words, and then learns embeddings for the words such that using the inner-product between the corresponding
embeddings we can approximate the logarithm of the co-occurrence counts between the words.  

For consistency of the comparison, we train all word embedding learning methods on the same ukWaC corpus\footnote{\url{http://wacky.sslmit.unibo.it/doku.php?id=corpora}} which is a web-derived corpus of English consisting of ca. 2 billion words \citep{ferraresi2008introducing}.
We lowercase all the text and tokenise using NLTK\footnote{\url{http://www.nltk.org/_modules/nltk/tokenize.html}}. 
We use the publicly available implementations by the original authors of CBOW, SG, and GloVe for training the word embeddings using the recommended parameters settings. 
Specifically, the context window is set to 5 words before and after the target word, and words with frequency less than 6 in the corpus are ignored, resulting in a vocabulary 
containing  1,371,950 unique words.
The negative sampling rate in SG is set to 5 words for each co-occurrence.
Our vocabulary is restricted to the words that appeared more that 6 times in the corpus, resulting in a vocabulary which includes 1,371,950 unique words. 
Using each of the word embedding learning methods, we learn 300 dimensional word embeddings.

In addition to prediction-based word embeddings described above (i.e. CBOW, SG, and GloVe), we evaluate counting-based word representations for relation representation. 
This method assigns each word with a high-dimensional vector that captures the contexts in which it occurs. We first construct unigram counts from the ukWaC corpus. The co-occurrences between low-frequency words are rare and result in a sparse co-occurrence matrix. To avoid this issue, we consider the most-frequent 50,000 words in the corpus as our vocabulary, and consider co-occurrences between only those words. 
We found that a vocabulary of 50,000 frequent words to be sufficient for covering all the benchmark datasets used in the evaluations. 
Moreover, truncating the co-occurrence matrix to the top frequent contexts makes the dimensionality reduction methods computationally inexpensive. Then the word-word co-occurrence statistics are computed from the corpus using windows of size 5 tokens on each side of the target word. 
We weight the co-occurrences by the inverse of the distance between the two words measured by the number of tokens that appear between the two words.
Afterwards, Positive Pointwise Mutual Information (PPMI) 
is computed from the co-occurrence matrix $\mat{W} \in \mathbb{R}^{m \times n} $ as follows:
\begin{equation}
\label{eq:ppmi}
\textrm{PPMI}(x,y)=\max \left( 0,\log \frac{p(x,y)}{p(x)p(y)} \right),
\end{equation}  
where $p(x,y)$ is the joint probability that the two words $x$ and $y$ co-occurring in a given context, whereas $p(x)$ is the marginal probability.
 We then apply Singular Value Decomposition (SVD) to the PPMI matrix, which factorises $\mat{W}$ as, $\mat{W}=\mat{U}\mat{S}\mat{V}\T$, where $\mat{S}$ is the singular values of 
 $\mat{W}$. We truncate $\mat{S}$ keeping only the top 300 singular values to reduce the dimensionality and thus increase the density of words representation. This count-based statistical method for word representations is widely applied in NLP to produce semantic representations for words and documents~\cite{clark2015vector,turney2010frequency}. 
 
 As an alternative dimensionality reduction method, we use Nonnegative Matrix Factorisation (NMF) in our experiments~\citep{lee2001algorithms}.  
 Given a matrix $\mat{W} \in \R^{m \times n}$, NMF computes the factorisation $\mat{W} = \mat{G} \mat{H}$, where $\mat{G} \in \R^{m \times d}$,
 and $\mat{H} \in \R^{d \times n}$, and $\mat{G} \geq 0, \mat{H} \geq 0$ (i.e. $\mat{G}$ and $\mat{H}$ contain non-negative elements).
 By setting $d < \min(n,m)$, we can obtain lower $d$-dimensional embeddings for the rows and columns of $\mat{W}$,
 given respectively by the rows and columns in $\mat{G}$ and $\mat{H}$.
 Unlike, SVD, the embeddings created using NMF are non-negative.
 By using non-negative  embeddings in our evaluations,
 we can test the behaviour of the different relation composition operators under nonnegativity constraints.

\section{Evaluation methods}
\label{sec:eval-methods}

Prior work that proposed compositional operators such as Mult, Add etc. evaluate their effectiveness on semantic composition tasks. For example, \citet{Mitchell:ACL:2008,Mitchell:COGSCI:2010} used a crowd sourced dataset of phrase similarity. First, a phrase is represented by applying a particular compositional operator on the constituent word representations. Next, the similarity between two phrases is computed using some similarity measure such as the cosine similarity between the corresponding phrase representations. Finally, the computed similarity scores are compared against human similarity ratings using some correlation measure such as the Spearman or Pearson correlation coefficients. If a particular compositional operator produces a higher agreement with human similarity ratings then it is considered superior. However, our task in this paper is to measure similarity between relations and not phrases (or sentences). Therefore, this evaluation protocol is not directly relevant to us. Instead, we use word analogy detection (Section~\ref{sec:relsim}) and knowledge base completion (Section~\ref{sec:KBCompletion}) tasks, which are more dependent on better relation representations.  
 
\subsection{Relational similarity prediction}
\label{sec:relsim}

Given two word pairs $(a,b)$ and $(c,d)$, the task is to measure the similarity between the semantic relations that exist between the two words in each pair.
This type of similarity is often referred to as \emph{relational similarity} in prior work~\cite{turney2006similarity}.
The task is to measure the degree of relational similarity between two given word-pairs $(a,b)$ and $(c,d)$. We need a method that assigns a high degree of relational similarity if the first pair stands in the same relation as another pair. Two benchmark datasets have been used
frequently in prior work for evaluating relational similarity measures are  SAT \citep{turney2003combining} dataset
and SemEval 2012-Task2\footnote{\url{https://sites.google.com/site/semeval2012task2/}} \citep{jurgens2012semeval} dataset.
Next, we briefly describe the protocol for evaluating relational similarity measures using those datasets. 

The Scholastic Aptitude Test (SAT) word analogy dataset contains 374 multiple choice questions in which each question contains a word-pair as the stem, and the examinees are required to select the most analogous word-pair from a list of 4 or 5 candidate answer word-pairs.
An example is shown in Table \ref{table:SatExample}. We generate relation embeddings for the question and its choice word-pairs using a compositional operator.
Next, the cosine similarity (Equation \ref{equ:cos}) between the relation representation $\vec{x}$ of the question word-pair $(a,b)$ and the relation representation $\vec{y}$ of each of the candidate word-pairs $(c,d)$ is computed to select the candidate with the highest  similarity score as the  correct answer. Cosine similarity between two vectors is defined as follows:
\begin{equation}
\label{equ:cos}
\textrm{sim}(\vec{x},\vec{y})=\textrm{cos}(\theta)= \frac{\vec{x}\T\vec{y}} {\norm{\vec{x}} \norm{\vec{y}}}
\end{equation}
The recorded accuracy is the ratio of the number of questions that are answered correctly to the total number of the questions in the dataset.
Because there are five candidate answers out of which only one is correct, random guessing would give a 20\% accuracy. 

\begin{table}[t]
\centering 
\scalebox{0.8}{
\begin{tabular}{ c c c } 
 \hline
 Stem: & &  ostrich:bird\\
  \hline
 Choices: & (a) & lion:cat \\ 
  & (b) & goose:flock\\ 
  & (c) & ewe:sheep \\ 
  & (d) &cub:bear \\ 
  & (e) &primate:monkey\\ 
  \hline 
  Solution: & (a) & lion:cat \\ 
  \hline 
\end{tabular}}
\caption{An example question from the SAT dataset. In this question, the common relation between the stem (ostrich, bird) and the correct answer (lion, cat) is \textsf{is-a-large}.}
\label{table:SatExample}
\end{table}

SemEval 2012 Task-2 covers 10 categories of semantic relations, each with a number of subcategories.
In total the dataset has 79 subcategories. Each subcategory (relation) has approximately 41 word pairs and three to four prototypical examples. 
Example word-pairs from the SemEval dataset are illustrated in Table \ref{table:SemEvalExample}. 
The task here is to assign a score to each word-pair, which indicates the average of the relational similarity between the given word-pair and prototypical word-pairs in a subcategory.

\begin{table}[t]
\centering 
\small
\scalebox{0.65}{
\begin{tabular}{ c c c c } 
 \hline
 Main Category & Description & Subcategories&  Prototypical pairs\\
  \hline
PART-WHOLE & One word names a part of the & \emph{Object:Component} & car:engin, face:nose  \\ 
& entity named by the other word & \emph{Mass:Portion} & water:drop, time:moment \\
& &  \emph{Collection:Member}&  forest:tree, anthology:poem \\
\hline 
CLASS-INCLUSION & One word names a class that includes & \emph{Taxonomic} & flower:tulip, poem:sonnet  \\ 
&  the entity named by the other word & \emph{Functional} & weapon:knife, ornament:brooch \\
& &  \emph{Class Individual}& river:Nile, city:Berlin \\  
\hline 
CAUSE-PURPOSE & One word represents the cause, purpose or goal & \emph{Cause:Effect} & enigma:puzzlement, joke:laughter  \\ 
&  of entity named by the other word, or the purpose & \emph{Case:Compensatory Action} & hunger:eat, fatigue:sleep \\
& or goal of using the entity named by the other wod&  \emph{Enabling Agent:Object}& match:candle, gasoline:car \\  
\hline 
\end{tabular}}
\caption{Example of taxonomy of the semantic relations in SemEval dataset}
\label{table:SemEvalExample}
\end{table}

An alternative approach for measuring the accuracy of a relation embedding method is to apply the relation embedding to complete word analogies.
measuring relational similarity could be evaluated in terms of completing an analogy $a:b :: c:?$. 
In other words, we must find the fourth (missing) word $d$ from a fixed vocabulary such that the relational similarity between $(a,b)$ and $(c,d)$ is maximised.
Equation \ref{eq:analogyques} uses the \textsf{PairDiff} operator for representing the relation between two words, and use
cosine similarity to measure the relational similarity between the two word-pairs. 
Likewise, we can use the other compositional operators \textsf{Add} and \textsf{Mult} to first create a relational embedding and then use cosine similarity to measure relational similarity. 

For the analogy completion task  we use two  datasets: MSR \citep{mikolov2013linguistic}, and Google analogy \citep{Mikolov:NIPS:2013} datasets. 
MSR dataset contains 8,000 proportional analogies covering 10 different syntactic relations, whereas the Google contains 19,544 analogical word-pairs covering 9 syntactic and 4 semantic relation types, corresponding to 10,675 syntactic and 8,869 semantic analogies. 
We restrict the search space for the missing word to the words that appear in a large set of vocabulary consists of 13,609 words in ukWaC, excluding the three words for each question.  
\begin{equation}
d^*= \arg \max_{d \in V} (\textrm{cos}(\vec{v}_b - \vec{v}_a, \vec{v}_d - \vec{v}_c))
\label{eq:analogyques}
\end{equation}

\subsection{Relation classification}
\label{sec:classification}

In relation classification, the problem is to classify a given pair of words $(w_1,w_2)$ to a specific relation $r$ in a predefined set of relations $\cR$ according to the relation that exists between $w_1$ and $w_2$ . We use the  DiffVecs dataset proposed by \citet{vylomova2015take} that consists of 12,458 triples $\langle w_1,w_2,r \rangle$, where word  $w_1$ and $w_2$ are connected by a relation $r$. 
The relation set $\cR$ includes 15 relation types comprising lexical semantic relations, morphosyntactic paradigm relations and morphosemantic relations.\footnote{https://github.com/ivri/DiffVec}

We use the different compositional operators discussed in Section~\ref{sec:operators} to represent each word-pair by a relational embedding.
We then perform 1-nearest neighbour (1-NN) classification in this relational embedding space to classify the test word-pairs into the relation types. 
 If the nearest neighbour has the same relation label as the target word-pair, then we consider it to be a correct classification. 
 The classification accuracy is computed as follows: 
\begin{equation}
\small
\label{eq:salience}
\textrm{Accuracy} = \frac{\textrm{correct matches}}{\textrm{total number of pairs}}
\end{equation}

We experimented using both unnormalised word embeddings as well as $\ell_{2}$ normalised word embeddings.
We found that $\ell_{2}$ normalised word embeddings perform better than the unnormalised version in most configurations.
Consequently, we report results obtained only with the $\ell_{2}$ normalised word embeddings in the remainder of the paper.

\subsection{Knowledge base completion}
\label{sec:KBCompletion}
Knowledge graphs such as WordNet and FreeBase that link entities according to numerous relation types that hold between entities are important resources for numerous NLP tasks such as question answering, entity and relation extraction.
Automatic knowledge base completion attempts to overcome the incompleteness of such knowledge bases by predicting missing relations in a knowledge base.
For instance, given a first entity (also known as the head entity $h$) and a relation type $r$, we need to predict a second entity (also known as the tail entity $t$) such that $h$ and $t$ are related by $r$. 

To evaluate the unsupervised compositional operators for the knowledge base completion task, we apply the following procedure. 
First, we require pre-trained entity embeddings as the input to a compositional operator.
Translating embeddings (TransE) model is one of the popular methods  for learning entity representations from a given knowledge graph~\cite{bordes2013translating}. 
In TransE, if $(h,r,t)$ holds, then the entity embeddings are learnt such that: $\vec{h}+\vec{r}\approx \vec{t}$.
We consider two knowledge bases frequently used in prior work on knowledge base completion~\cite{bordes2013translating}.
Namely, WordNet (WN18) and FreeBase (FB15k). The datasets and the source code that generates entity embeddings are publicly available~\cite{lin2015learning}\footnote{https://github.com/thunlp/KB2E}. 

To evaluate the accuracy of a relation composition operator $f$, we first create a representation $\vec{r}_{i}$ for each relation type $r$ using the entity pairs$(h,t)$  in the training data by applying $f$ to the embeddings of the two entities $h$ and $t$ as follows: 
\begin{equation}
\label{eq:KB-r}
\vec{r}=\frac{1}{|\mathcal{R}|}\sum_{(h,r,t) \in \mathcal{R}} f(h,t)
\end{equation}
Here, $\mathcal{R}$ is the set of pairs of entities that are related by $r_i$.
 
Next, for each test triple $(h',r',t')$, we rank the candidate tail entities $t'$ according to the cosine similarity between each of the
relation embedding $\vec{r}'$ of the relation $r'$ computed using \eqref{eq:KB-r}, and the result of applying $f$ to the entity embeddings
$\vec{h}'$ and $\vec{t}'$. The cosine similarity score we used to rank candidate tail entities is given by,
\begin{equation}
\label{eq:KB-cos}
cos(\vec{r},f(h',t')).
\end{equation}
We rank all tail entities in all test entity pairs according to \eqref{eq:KB-cos} and select the top-ranked entity as the correct completion.
This process is repeatedly applied for predicting the head entities for each test triple as well.

If the correct tail (or head) entity (according to the original test tuple) can be accurately predicted using the relation embeddings
created by applying a particular compositional operator, then we can conclude that operator to be accurately capturing the relational information.
Following prior work on knowledge base completion, we use two measures for evaluating the predicted tail (or head) entities: Mean Rank and Hits$@$10. Mean rank is the average rank assigned to the correct tail (or head) entity in the ranked listed of candidate entities according to \eqref{eq:KB-cos}. A lower mean rank is better because the correct candidate is ranked at the top by the compositional operator under evaluation.
Hits$@$10 is the proportion of correct entities that have been ranked among the top 10 candidates. 
It is noteworthy that our purpose here is not to propose state-of-the-art knowledge base completion methods.
We are using knowledge base completion simply as an evaluation task to compare different compositional operators, whereas the prior works 
in knowledge base completion learn entity and relation embeddings that can accurately predict the missing relations in a knowledge base. 

\section{Experimental results} 
\label{ExperRes}

\subsection{Performance of Relational Similarity Task}
\label{sec:operator-performance}
In Table \ref{tab:accuracy}, we compare the four compositional operators (PairDiff, Concat, Add and Mult) described in Section~\ref{sec:operators} for the four different word representation models as described in Section~\ref{WRM}. 
We observe that PairDiff achieves the best results compared with other operators for all the evaluated datasets and all the word representation methods. 
PairDiff is significantly better than Add or Mult for all embeddings (both prediction- and counting-based) in MSR, Google and DiffVec datasets according to Clopper-Pearson confidence intervals (p $<$ 0.05).
SAT is the smallest dataset among all, so we were unable to see any significant differences on SAT.

Analogy completion in Google and MSR datasets are considered as an open vocabulary task because to answer a question of the form \enquote {\emph{a} is to \emph{b} as \emph{c} is to ?}, we must consider all the words in the corpus as candidates, which is an open vocabulary, not limited to the words that appear in the benchmark datasets as in SAT or SemEval datasets.
Therefore, applying PairDiff to each pair $(a,b)$ and $(c,d)$ will retrieve candidates \emph{d} that have relations with \emph{c} similar to the relation between \emph{a} and \emph{b}, but not necessary similar to the word \emph{c}. For instance, the top 3 ranked candidates for a question \enquote {\emph{man} is to \emph{woman} as \emph{king} is to ?} are \emph{women}, \emph{pregnant} and \emph{maternity}. We notice that the top ranked candidates indicate feminine entities. This explains the performance of PairDiff on MSR and Google datasets, which is lower compared with other datasets. 
Similar observations have been made by \citet{levy2014linguistic}. 
Moreover, the open vocabulary task (Google and MSR) is harder than the closed vocabulary task (SAT, SemEval and DiffVecs) as the number of incorrect candidates is much larger in the open vocabulary setting. 
This means that the probability of accidentally retrieving a noisy negative candidate as the correct answer is higher than in the closed vocabulary task. 
 \begin{center}
\begin{table}[h]
\centering
\small
\scalebox{0.8}{
\begin{tabular}{| c | c| c| c| c| c| c| c | c|} 

 \hline
 Representation   & Compositional   & SAT & SemEval &MSR& \multicolumn{3}{c|}{Google}   &  D{\small IFF}V{\small ECS}\\
 \cline{6-8}
 model& operator &&&&Sem.&Syn.&Total &\\
 \hline\hline
 CBOW & PairDiff  &   \textbf{41.82} 	& \textbf{44.35}   &	\textbf{30.16 }&  \textbf{ 24.43}	  &   \textbf{32.31 } 	&    \textbf{ 28.74} &\textbf{ 87.38}\\  \cline{2-9}

 & Concat &   38.07	&  41.06    &  0.39  &	  3.01  &    1.26 &   2.05  &  83.74 \\  \cline{2-9}
 
 & Add &  31.1    &  36.37   & 0.06  &	0.16   &   0.15	 & 0.15   &  79.27\\  \cline{2-9}

&Mult&  27.88   &   35.19  	& 8.13 &   2.38	&   6.11 	&  4.42  & 79.16\\ 

\specialrule{1pt}{1pt}{1pt}

SG & PairDiff  &  39.41  	&   44.03	& 21.08 & 22.28	  &  26.47  	&    24.57 &86.32\\  \cline{2-9}

 & Concat &   35.92   &      41.21	&  0.3  &	1.4    &    1.17	 &  1.27   & 81.19  \\  \cline{2-9}
 
 & Add &     28.69	&  35.48  	& 0.0& 0.17	  &  0.13	 &  0.15  &78.48\\  \cline{2-9}

&Mult&   24.4 	&    35.4	&3.26 &  2.29	&  4.47 	&  3.48 & 78.33\\ 

\specialrule{1pt}{1pt}{1pt}

GloVe & PairDiff  & 41.02   & 42.8  &  16.74&15.42 	& 21.0   & 18.47 & 83.87\\ \cline{2-9}

 & Concat &   36.19   &      40.17	&  0.31  &	  2.27  &    1.17	 &   1.67  &  81.1 \\  \cline{2-9}
 
& Add  & 29.22   & 35.23  &0.0 &  0.24	& 0.18   &   0.2 &73.76\\ \cline{2-9}

&Mult&  23.32 	&  32.0	&0.91 & 3.87	& 1.39 	&  2.51 &66.32\\  

\specialrule{1pt}{1pt}{1pt}

SVD & PairDiff  &  36.9  	&  43.44  &8.49 &  2.84 	  &   11.26 	&  7.44     & 85.8 \\  \cline{2-9}

 & Concat &    38.77   &   42.04   &  0.35   &	0.5     &   0.82   &  0.68    &  81.25  \\  \cline{2-9}

& Add &      31.82 	&  36.05    	&   0.01 &	  0.26  &   0.14 &  0.19   & 77.93  \\  \cline{2-9}

&Mult&   29.14   &    34.79 &   5.56 &    0.52	&    6.91 	&   4.01  & 77.58 \\ 

\specialrule{1pt}{1pt}{1pt}

NMF & PairDiff  &   35.29 	&  42.88  & 2.8 &  1.75 	  & 3.66  	&  2.79    & 84.66\\  \cline{2-9}

 & Concat &   31.02   &      41.39	&  0.19  &	  0.44  &    0.65	 &   0.5  &  81.4 \\  \cline{2-9}

& Add &   29.68  &    36 &   0.03   &   0.21	 &   0.11 &0.16&77.56  \\  \cline{2-9}

&Mult&   21.12  &   34.49  	& 0.0 &   0.03	&    0.0	& 0.02   &56.99 \\ 

\hline

\end{tabular}}
\caption{Accuracy of the compositional operators for relational similarity prediction and relational classification (last right column). }   \label{tab:accuracy}
\end{table}
\end{center}

Mult is performing slightly worse with NMF compared to other embeddings. 
Recall that NMF produces non-negative embeddings and Mult is performing an elementwise multiplication operation on the two input word embeddings
to create the embeddings for their relation. If the negativity was the only issue with Mult operator as previously suggested by \cite{Baroni:EMNLP:2010},
then Mult should have performed better with NMF. We hypothesise the issue here is sparsity in the relation representations.
To test our hypothesis empirically we conduct the following experiment.

First, we randomly select 140 word-pairs from the Google dataset and apply different compositional operators to create relation embeddings for
each word-pair using 300 dimensional CBOW word embeddings as the input.
Next, we measure the average sparsity of the set of relational embeddings created by each operator. 
We define sparsity at a particular cut-off level $\epsilon$ for a $d$ dimensional vector as the percentage of elements with absolute value less than
or equal to $\epsilon$ out of $d$. Formally, sparsity is given by \eqref{eq:sparsity}.
\begin{equation}
 \label{eq:sparsity}
 \textrm{sparsity} = \frac{1}{d}\sum_{i=1}^{d} \cI [|x_{i}| \leq \epsilon]
\end{equation}
Here, $\cI$ is the indicator function which returns 1 if the expression evaluated is true, or 0 otherwise.
Our definition of sparsity is a generalisation of the $\ell_{0}$ norm that counts the number of non-zero elements in a vector.
However, in practice, exact zeros will be rare and we need a more sensitive measure of sparsity, such as the one given in \eqref{eq:sparsity}.
Average sparsity is computed by dividing the sum of sparsity values given by \eqref{eq:sparsity} for the set of word-pairs by the number of
word-pairs in the set (i.e. 140).

\begin{figure}[h]
\centering
\includegraphics[scale=0.4]{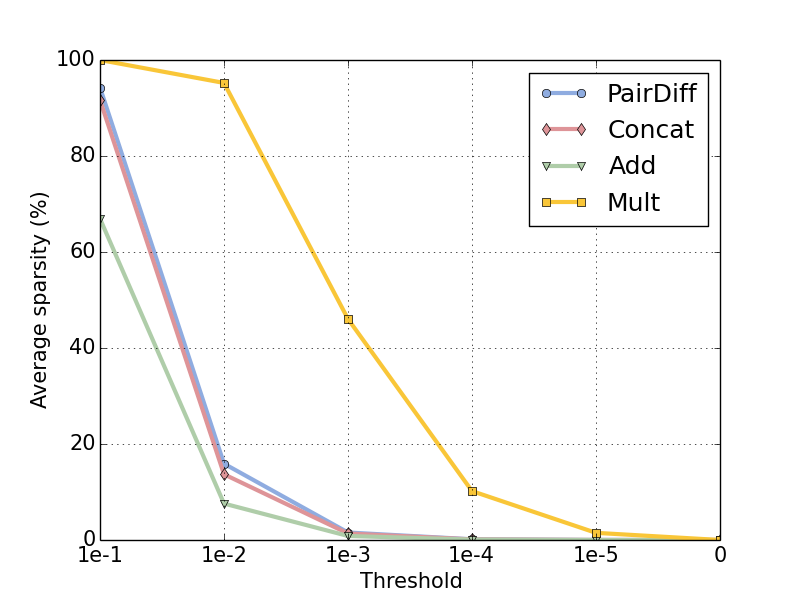}
\caption{The average sparsity of relation embeddings for different operators using CBOW embeddings with 300 dimensions for some selected pairs of words.}
\label{fig:sparsity}
\end{figure}

Figure~\ref{fig:sparsity} shows the average sparsity values for different operators under different $\epsilon$ levels.
Figure \ref{fig:sparsity} shows that Mult operator generates sparse vectors for relations compared to other operators under 
all $\epsilon$ values. Considering that Mult is performing a conjunction over the two input word embeddings, even if at least one embedding
has a nearly zero dimension, after elementwise multiplication we are likely to be left with nearly zero dimensions in the relation embedding.
Such sparse representations become problematic when measuring cosine similarity between relation embeddings, which leads to poor performances
in word analogy tasks.

\subsection{Effect of Dimensionality}
\label{sec:dim}

The dimensionality of the relational embeddings produced by the compositional operators presented in Section~\ref{sec:operators} depends on the dimensionality of the input word embeddings. For example, Mult, Add, and PairDiff operators produce relational embeddings with the same dimensionality as the input word embeddings, whereas the Concat operator produce relational embeddings twice the dimensionality of the input word embedding. 
A natural questions therefore is that how does the performance of the relational embeddings vary with the dimensionality of the input word embedding.
To study the relationship between the dimensionality of the input word embedding and the composed relational embedding we conduct the following experiment. 

\begin{figure}[h!]
\centering
\includegraphics[scale=0.48]{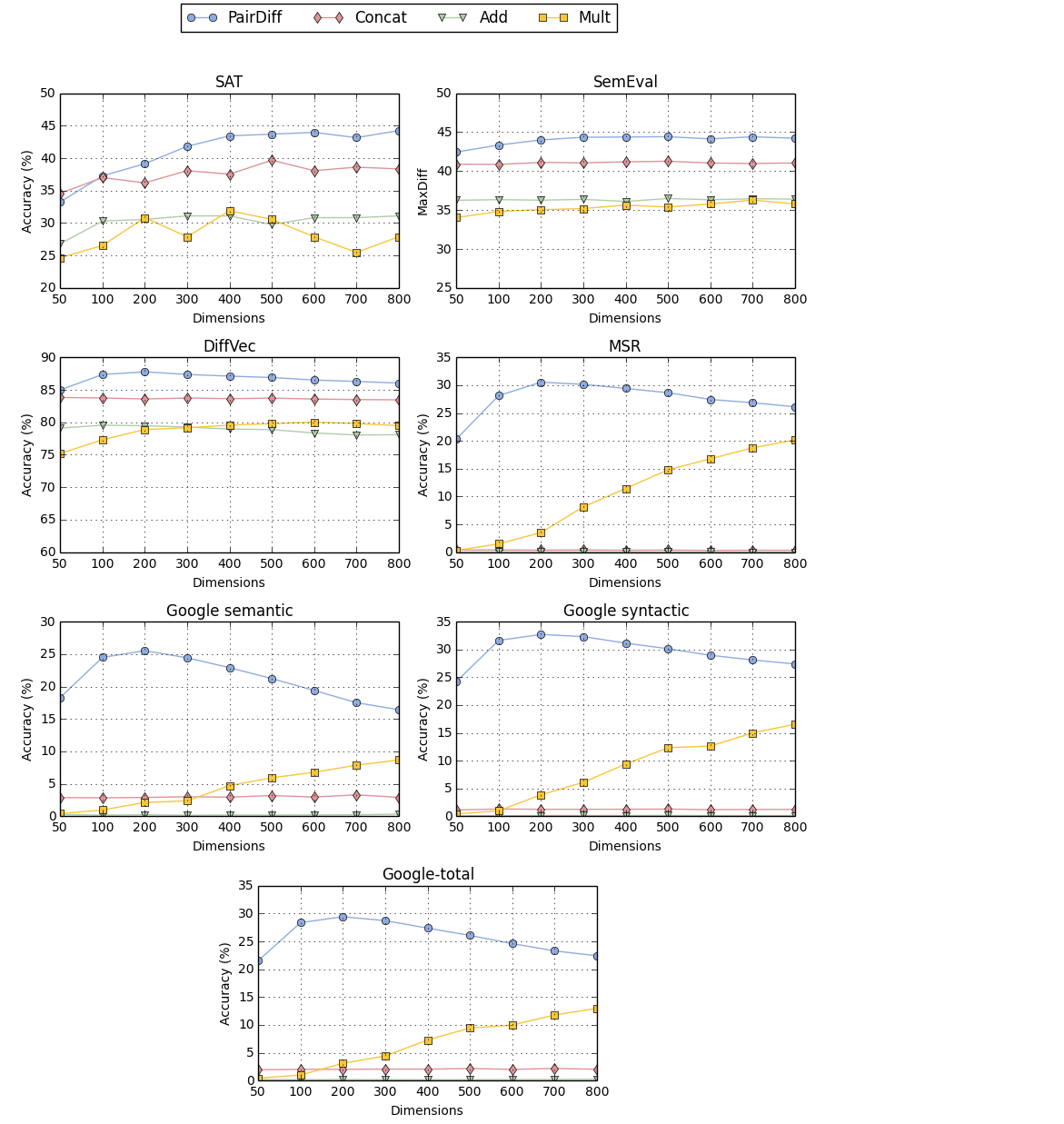}
\caption{The effect of the dimensionality of the CBOW word embeddings for compositional relation representations.}
\label{fig:Dimensionality}

\end{figure}

We first train word embeddings of different dimensionalities using the ukWaC corpus. 
We keep all the other parameters of the word embedding learning method fixed except for the dimensionality of the word embeddings learnt. 
Because CBOW turned out be the single best word embedding learning method according to the results in Table~\ref{tab:accuracy},
we use CBOW as the preffered word embedding learning method in this analysis.
Figure \ref{fig:Dimensionality} shows the performance of the different compositional operators on the benchmark datasets using CBOW input word embeddings with dimensionalities in the range 50-800.

As seen from Figure~\ref{fig:Dimensionality}, PairDiff outperforms all other operators across all dimensionalities.
The best results on SemEval and DiffVecs datasets are reported by PairDiff with 200 dimensions.
Performance saturates when the dimensionality is increased beyond this point. 
On the other hand, SAT shows different trend.
On SAT, the performance of PairDiff continuously increases with the dimensionality of the input word embeddings.
On the other hand,  in MSR and Google datasets we see a different trend where
the performance of PairDiff decreases while that of Mult increases with the dimensionality of the input word embedding.

To understand the above-described trends first note that the dimensions in word embeddings are providing almost complementary
 information related to the semantics of a word.\footnote{As described in Section~\ref{sec:operators}, Pearson correlation coefficients
 between different dimensions in word embeddings are small, showing that different dimensions are uncorrelated.} 
Adding more dimensions to the word embedding can be seen as a way of representing richer semantic information.
However, increasing the dimensionality also increases the number of parameters that we must learn.
Prediction-based word embedding learning methods first randomly initialise all the parameters and then update them such that the
co-occurrences between words can be accurately predicted in a given context window.
However, the training dataset, which in our case is the ukWaC corpus, is fixed.
Therefore, we will have more parameters than we could reliably estimate using the data we have, resulting in some
overfitted noisy dimensions as we increase the dimensionality of the word embeddings learnt.

One hypothesis for explaining the seemingly contradictory behaviour with PairDiff and Mult operators is as follows.
When we increase the dimensionality of the input word embeddings, there will be some noisy dimensions in the input word embeddings.
PairDiff operators amplifies the noise in the sense that the resultant offset vector will retain noisy high dimensions that appear in both
word embeddings. 
On the other hand, Mult operator can be seen as a low-pass filter where we shutdown dimensions that have small (or zero) valued dimensions
in at least one of the two embeddings via the element-wise multiplication of corresponding dimensions.
Therefore, Mult will be robust against the noise that exist in the higher dimensions of the word embeddings than the PairDiff operator.

\begin{figure}[h]
\centering
\includegraphics[scale=0.35]{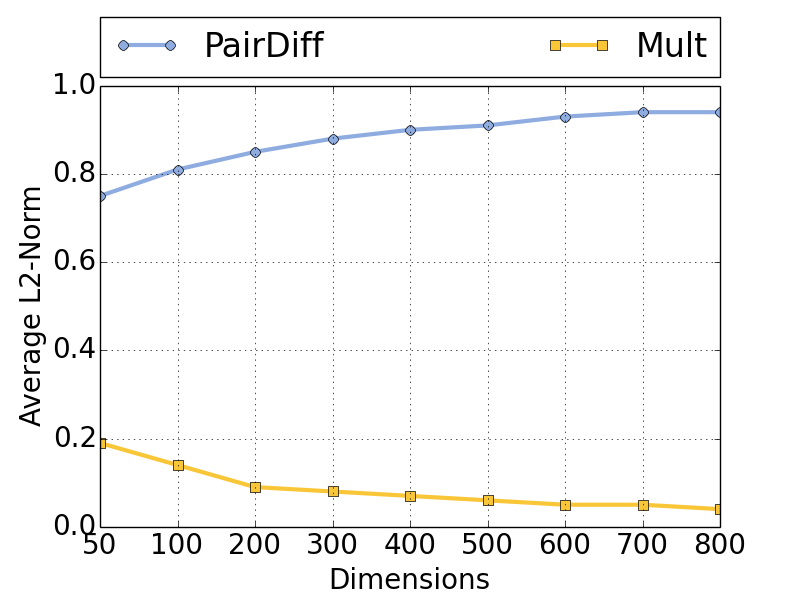}
\caption{Average $\ell_2$ norm of relational vectors generated using PairDiff and Mult operators.}
\label{fig:AverageNorm}
\end{figure}

To empirically test this hypothesis we compute the $\ell_{2}$ norm of $(\vec{v}_a - \vec{v}_b)$ and $(\vec{v}_a \odot \vec{v}_b)$ for
word embeddings of different dimensionalities and compute the average over 140 randomly selected word-pairs. 
As shown in Figure \ref{fig:AverageNorm}, the norm of PairDiff relation embedding is increasing with dimensionality, 
whereas norm of the relation embedding generated by Mult decreases.
This proves our hypothesis that Mult is filers out the noise in high dimensional word embeddings better than PairDiff.
 
\subsection{Performance on Knowledge Base Completion Task}

Table~\ref{tab:KBtask} displays the performance on the compositional operators for the knowledge base completion task on the two knowledge graphs WN18 and FB15k, where low mean rank and high Hits$@$10 indicates better performance. 
As can be seen from the Table, Mult operator yields the lowest mean rank and the highest Hits$@$10 accuracy among other operators for the both knowledge bases. 

Given that PairDiff was the best operator for relational similarity tasks, it is surprising that Mult operator outperforms PairDiff in both WN18 and FB15k datasets.
Recall that knowledge base completion is the task where given that $(h,t)$ pair is related by a relation $r$ (as provided in the train set)
we need to assess how likely that $(h\textprime,t\textprime)$ is related by $r$.
In our evaluation process, this task is answered by measuring the inner-product between $f(h,t)$ and $f(\vec{h}\textprime,\vec{t}\textprime)$, where $f$ is a compositional function that represents the relationship between $h$ and $t$. 
In the case of Mult operator, we have: 
$\textrm{similarity-score}=(\vec{h}\odot \vec{t})^\top(\vec{h}\textprime \odot \vec{t}\textprime)$, this indicates that if a dimension is not common across all four entities it does not contribute to the overall similarity score. 
This can be seen as a strict way of estimating relational similarity between train and test pairs because a particular dimension must be \emph{on} in all four words involved in an analogy. 

On the other hand, PairDiff operator scores test entity pairs by $(\vec{h} - \vec{t})\T(\vec{h}' - \vec{t}')$.
Here, $(h',t')$ is an entity pair in the test dataset with the target relation $r$, and we are interested in finding, for example, candidate tail entities
$t'$ that has $r$ with a given head entity $h'$. This score can be further expanded as 
$(\vec{h} - \vec{t})\T\vec{h}' - (\vec{h} - \vec{t})\T\vec{t}'$. The first term is fixed given the training dataset and the head entity, and
the rank of the tail entity is determined purely based on  $(\vec{h} - \vec{t})\T\vec{t}'$, where the head entity $h'$ does not participate in.
This is problematic because entities $t'$ that are similar to $t$ and dissimilar to $h$ will be simply ranked at the top irrespective of the relation
$t'$ has with $h'$. Indeed in Table~\ref{tab:KBtask} we see that mean rank for PairDiff is significantly higher compared to that of Mult.
This suggests that many irrelevant tail (or head) entities are ranked ahead of the correct entity for each test tuple.
On the other hand, in relational similarity task, the two pairs between which we must measure similarity are fixed and this issue is not

If a relation is asymmetric such as hypernym and hyponym as in WN18, addition model will be insensitive to the directionality of such relations compared to PairDiff which explains the better performance of PairDiff over Add.
 \begin{center}
\begin{table}[h]
\centering
\small
\scalebox{0.8}{
\begin{tabular}{| c| c| c| c| c|} 

 \hline
Compositional   &  \multicolumn{2}{c|}{WN18}   &  \multicolumn{2}{c|}{FB15k}\\
 \cline{2-5}
 operator& MeanRank & Hits$@$10(\%)&MeanRank&Hits$@$10(\%) \\
\specialrule{1pt}{1pt}{1pt}
 PairDiff & 13,198& 11.34&1,206 &44.4 \\
 Concat &9,896 &2.77 & 542&29.49 \\
 Add &12,178 &1.88 &1,211 &21.7 \\
 Mult &\textbf{812 }&\textbf{54.93 }&\textbf{ 256}&\textbf{ 50.66}\\
 \hline
\end{tabular}}
\caption{Accuracy of the compositional operators for knowledge base completion task. }   \label{tab:KBtask}
\end{table}
\end{center}

\subsection{Evaluating the Asymmetry of the PairDiff Operator}
\label{sec:direction}

Relations between words can be categorised as either being \emph{symmetric} or \emph{asymmetric}. 
If two words $a$ and $b$ are related by a symmetric relation $r$, then $b$ is also related to $a$ with the same relation $r$. 
Examples of symmetric relations include synonyms and antonyms. 
On the other hand, if $a$ is related to $b$ by an \emph{asymmetric} relation, then $b$ might not be necessarily related to $a$ with the same relation $r$. 
Examples of asymmetric relations include hypernyms and meronyms. 
As discussed in Section~\ref{sec:operator-performance}, \textsf{PairDiff} operator outperforms Add and Mult operators.
Unlike Mult and Add, which are commutative operators, PairDiff is a non-commutative operator. 
Therefore, PairDiff should be able to detect the direction of a relation. 

To test the ability of PairDiff to detect the direction of a relation, we set up the following experiment.
Using a set of word-pairs where there is a common directional relation $r$ between the two words in each word pair as training data,
we use PairDiff to represent the relationship between two words in a word-pair, given the word embeddings for those two words.
Next, we swap the two words in each word-pair and apply the same procedure to create relation embeddings for the reversed relation $r'$ in each word-pair.
We model the task of predicting whether a given word-pair contains the original relation $r$ or its reversed version $r'$ as a binary classification task.
Specifically, we train a binary support vector machine with a linear kernel with the cost parameter set to $1$ using held-out data.
If the trained binary classifier can correctly predict the direction of a relation in a word-pair, then we can conclude that the relation embedding for
that word-pair accurately captures the information about the direction of the relation that exists between the two words in the word-pair.
We can repeat this experiment with symmetric as well as asymmetric relation types $r$ and compare the performances of the trained classifiers
to understand how well the directionality in asymmetric relations is preserved in the PairDiff embeddings.

\begin{figure}[h]
\centering
\includegraphics[scale=0.5]{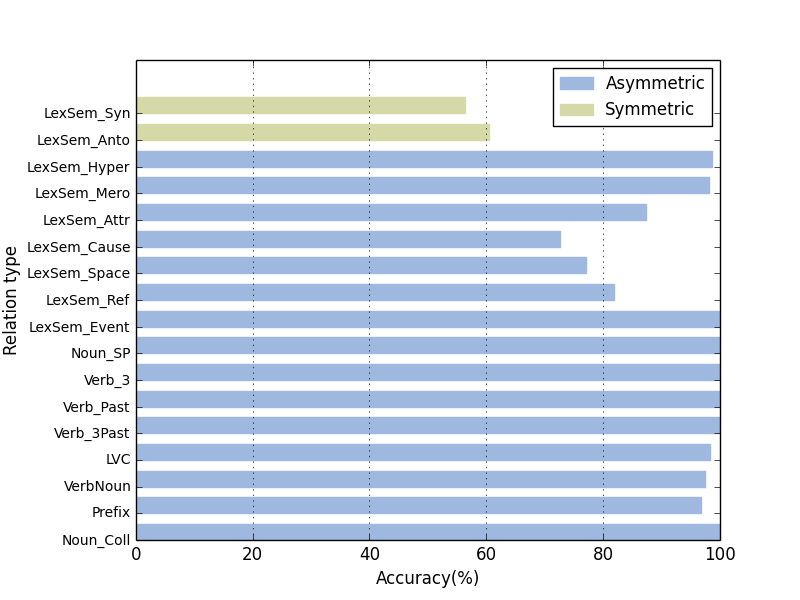}
\caption{The accuracy of SVM classifier.}
\label{fig:classification}
\end{figure}

For the asymmetric relation types we use all relation types in the D{\small IFF}V{\small ECS} because this dataset contains
only asymmetric relation types.
For symmetric relation types we use two popular symmetric semantic relations namely, synonymy\footnote{\url{http://saifmohammad.com/WebDocs/LC-data/syns.txt}} and antonymy\footnote{\url{http://saifmohammad.com/WebDocs/LC-data/opps.txt}}. 
We report five-fold cross-validation accuracies with each relation type in Figure \ref{fig:classification}. 
If the classifier reports a high classification accuracy for asymmetric relations than symmetric relations, then it indicates that the relation embedding
can encode the directional information in a relation.
From Figure~\ref{fig:classification} we see that, overall, the accuracies for the two symmetric relation types is lower than that for the
asymmetric relation types. 
This result indicates that PairDiff can correctly detect the direction in the asymmetric relation types.

\section{Discussion and conclusion}

This work evaluated the contribution of word embeddings for representing relations between pairs of words. 
Specifically, we considered several compositional operators such as PairDiff, Mult, Add, and Concat for creating a representation (embedding)
for the relation that exist between two words, given their word embeddings as the input. 
We used different pre-trained word embeddings and evaluated the performance of the operators on two tasks: relational similarity measurement
and knowledge base completion.
We observed that PairDiff to be the best operator for relational similarity measurement task, whereas Mult operator to be the
best for knowledge base completion task. We then studied the effect of dimensionality on the performance of these two operators and
showed that the sparsity of the input embeddings is affecting the Mult operator, and not the negativity of the input word embedding dimensions
as speculated in prior work.
Our analysis in this paper was limited to unsupervised operators in the sense that there are no parameters in the operators that can be 
(or must be) learnt from training data. 
This raises the question whether we can learn better compositional operators from labelled data to further improve the performance of
the compositional approaches for relation representation, which we plan to explore in our future work.

\bibliographystyle{elsarticle-num-names}
\bibliography{mybibfile}{}
 
\end{document}